\documentclass[a4paper,twoside]{article}
\usepackage[acronym]{glossaries}
\usepackage[hidelinks]{hyperref}

\usepackage{epsfig}
\usepackage{subcaption}
\usepackage{calc}
\usepackage{amssymb}
\usepackage{amstext}
\usepackage{amsmath}
\usepackage{amsthm}
\usepackage{multicol}
\usepackage{pslatex}
\usepackage{apalike}
\usepackage{algorithm2e}
\usepackage[bottom]{footmisc}
\usepackage{float}
\usepackage{SCITEPRESS} 
\usepackage{adjustbox} 

\usepackage[acronym]{glossaries}
\makeglossaries
\newacronym{ac}{AC}{Alternating Current}
\newacronym{acopf}{ACOPF}{Alternating Current Optimal Power Flow}
\newacronym{ai}{AI}{Artificial Intelligence}
\newacronym{dc}{DC}{Direct Current}
\newacronym{drl}{DRL}{Deep Reinforcement Learning}
\newacronym{dcopf}{DCOPF}{Direct Current Optimal Power Flow approximation}
\newacronym{ddpg}{DDPG}{Deep Deterministic Policy Gradient}
\newacronym{dn}{DN}{Distribution Network}
\newacronym{dl}{DL}{Deep Learning}
\newacronym{dnn}{DNN}{Deep Neural Networks}
\newacronym{dqn}{DQN}{Deep Q-learning}
\newacronym{ddqn}{DDQN}{Double Q-learning}
\newacronym{gnn}{GNN}{Graph Neural Networks}
\newacronym{gae}{GAE}{Generalized Advantage Estimate}
\newacronym{il}{IL}{Imitation Learning}
\newacronym{ipopt}{IPOPT}{Interior Point Optimizer}
\newacronym{opf}{OPF}{Optimal Power Flow}
\newacronym{pi}{PI}{Policy Iteration}
\newacronym{pf}{PF}{Power Flow}
\newacronym{scop}{SOCP}{Second Order Cone Programming}
\newacronym{ppo}{PPO}{Proximal Policy Optimization}
\newacronym{mdp}{MDP}{Markov Decision Process}
\newacronym{ml}{ML}{Machine Learning}
\newacronym{mlp}{MLP}{Multilayer Perceptron}
\newacronym{mpnn}{MPNN}{Message Passing Neural Networks}
\newacronym{nn}{NN}{Neural Networks}
\newacronym{res}{RES}{Renewable Energy Sources}
\newacronym{rl}{RL}{Reinforcement Learning}
\newacronym{trpo}{TRPO}{Trust Region Policy Optimization}
\newacronym{tso}{TSO}{Transmission System Operator}
\newacronym{vi}{VI}{Value Iteration}
\begin{document}

\title{Proximal Policy Optimization with Graph Neural Networks for Optimal Power Flow \thanks{This is the author's version of the work. It is posted here for your personal use. Not for redistribution. The definitive version is published in SCITEPRESS Proceedings of the 14th International Conference on Data Science, Technology and Applications.}}
\author{
    \authorname{
    Ángela López-Cardona\sup{1}, 
    Guillermo Bernárdez\sup{3}, 
    Pere Barlet-Rose\sup{1,2}, 
    Albert Cabellos-Aparicio\sup{1,2}}
    \affiliation{\sup{1}Universitat Politècnica de Catalunya, Barcelona, Spain}
     \affiliation{\sup{2}Barcelona Neural Networking Center, Barcelona, Spain}
      \affiliation{\sup{3} UC Santa Barbara, California, US}
    \email{angela.lopez.cardona@upc.edu, guillermo\_bernardez@ucsb.edu, \{pbarlet, acabello\}@ac.upc.edu}
}


\abstract{\acrfull{opf} is a key research area within the power systems field that seeks the optimal operation point of electric power plants, and which needs to be solved every few minutes in real-world scenarios. However, due to the non-convex nature of power generation systems, there is not yet a fast, robust solution for the full \acrfull{acopf}.
In the last decades, power grids have evolved into a typical dynamic, non-linear and large-scale control system ---known as the power system---, so searching for better and faster \acrshort{acopf} solutions is becoming crucial. The appearance of \acrfull{gnn} has allowed the use of \acrfull{ml} algorithms on graph data, such as power networks. On the other hand, \acrfull{drl} is known for its proven ability to solve complex decision-making problems. Although solutions that use these two methods separately are beginning to appear in the literature, none has yet combined the advantages of both. We propose a novel architecture based on the \acrfull{ppo} algorithm with \acrlong{gnn} to solve the \acrlong{opf}. The objective is to design an architecture that learns how to solve the optimization problem and, at the same time, is able to generalize to unseen scenarios. We compare our solution with the \acrfull{dcopf} in terms of cost. We first trained our \acrshort{drl} agent on the IEEE 30 bus system and with it, we computed the \acrshort{opf} on that base network with topology changes.}

\onecolumn \maketitle \normalsize \setcounter{footnote}{0} \vfill

\section{\uppercase{Introduction}}
\label{sec:intro}
After several decades of development, power grids have transformed into a dynamic, non-linear, and large-scale control system, commonly referred to as the power system \cite{gnn_review}. Today, this power system is undergoing changes for various reasons. Firstly, the high penetration of \acrfull{res}, such as photovoltaic plants and wind farms, introduces fluctuations and intermittence to power systems. This generation is inherently unstable, influenced by several external factors like solar irradiation and wind velocity for solar and wind power, respectively \cite{opf_drl}. Concurrently, the integration of flexible sources (e.g., electric vehicles) brings about modifications to networks, including relay protection, bidirectional power flow, and voltage regulation \cite{gnn_review}. Lastly, emerging concepts like Demand Response—defined as the alterations in electricity usage by end-use customers from their typical consumption patterns in response to variations in electricity prices over time—affect the operational point within the electrical grid \cite{elec_book}. All these transformations render the optimization of production in power networks increasingly complex. In this context, \acrlong{opf} comprises a set of techniques aimed at identifying the optimal operating point by optimizing the power output of generators in power grids \cite{elec_book}.

The traditional approach to solving the \acrshort{opf} involves numerical methods \cite{opf_drl}, with \acrfull{ipopt} \cite{pandapower} being the most commonly employed. However, as networks grow increasingly complex, traditional methods struggle to converge due to their non-linear and non-convex characteristics \cite{opf_drl}. Nonlinear \acrshort{acopf} problems are often approximated using linearized \acrshort{dcopf} solutions to derive real power outcomes, where voltage angles and reactive power flows are eliminated through substitution (thus removing \acrfull{ac} electrical behavior). This approximation, however, becomes invalid under heavy loading conditions in power grids \cite{rivero1}. Additionally, the \acrshort{opf} problem is inherently non-convex because of the sinusoidal nature of electrical generation \cite{elec_book}. Alternative techniques seek to approximate the \acrshort{opf} solution by relaxing this non-convex constraint, employing methods such as \acrfull{scop} \cite{elec_book}. In daily operations that necessitate solving \acrshort{opf} within a minute every five minutes, \acrshort{tso} is compelled to depend on linear approximations. The solutions derived from these approximations tend to be inefficient, resulting in power wastage and the overproduction of hundreds of megatons of CO2-equivalent annually. Today, fifty years after the problem was first formulated, we still lack a fast, robust solution technique for the complete \acrlong{acopf} \cite{pf_hist}. For large and intricate power system networks with numerous variables and constraints, achieving the optimal solution for real-time \acrshort{opf} in a timely manner demands substantial computing power \cite{deep_opf}, which continues to pose a significant challenge.

In power systems, as in many other fields, algorithms of \acrshort{ml} have recently begun to be utilized. The latest proposals employ \acrlong{gnn}, a neural network that naturally facilitates the processing of graph data \cite{review_gnn_power}. An increasing number of tasks in power systems are being addressed with \acrshort{gnn}, including time series prediction of loads and \acrshort{res}, fault diagnosis, scenario generation, operational control, and more \cite{opf_gnn_cutre}. The primary advantage is that by treating power grids as graphs, \acrshort{gnn} can be trained on specific grid topologies and subsequently applied to different ones, thereby generalizing results \cite{review_gnn_power}. Conversely, \acrlong{drl} is recognized for its ability to tackle complex decision-making problems in a computationally efficient, scalable, and flexible manner—problems that would otherwise be numerically intractable \cite{opf_drl}. It is regarded as one of the state-of-the-art frameworks in Artificial Intelligence (AI) for addressing sequential decision-making challenges \cite{drl_gnn_rev}. The \acrshort{drl} based approach seeks to progressively learn how to optimize power flow in electrical networks and dynamically identify the optimal operating point. While some approaches utilize various \acrshort{drl} algorithms, none have integrated it with \acrshort{gnn}, which limits their ability to generalize and fully leverage the information regarding connections between buses and the properties of the electrical lines that connect them. Given this context, and considering that the combination of \acrshort{drl} and \acrshort{gnn} has demonstrated improvements in generalizability and reductions in computational complexity in other domains \cite{drl_gnn_rev}, we explore their implementation in this work.

\textbf{Contribution:} This paper presents a significant advancement through the proposal of a novel architecture that integrates the \acrlong{ppo} algorithm with \acrlong{gnn} to address the \acrlong{opf} problem. To the best of our knowledge, this unique architecture has not been previously applied to this challenge. Our objective is to rigorously test the design of our architecture, demonstrating its capability to solve the optimization problem by effectively learning the internal dynamics of the power network. Additionally, we aim to evaluate its ability to generalize to new scenarios that were not encountered during the training process. We compare our solution against the \acrshort{dcopf} in terms of cost, following the training of our \acrshort{drl} agent on the IEEE 30 bus system. Through various modifications to the base network, including changes in the number of edges and loads, our approach yields superior cost outcomes compared to the \acrshort{dcopf}, achieving a reduction in generation costs of up to 30\%.

\section{\uppercase{Related work}}
Until this paper, there had been no solution for the \acrshort{opf} problem that utilized \acrshort{gnn} to handle graph-type data and \acrshort{drl}, enabling generalization and understanding the internal dynamics of the power grid. Nevertheless, methods can be found in the literature that employ each of the approaches independently. Data-driven methods based on deep learning have been introduced to solve \acrshort{opf} in approaches such as \cite{rivero1}, \cite{balthazar1}, \cite{balthazar2}, \cite{deep_opf}, and \cite{opf_ml2}, among others. However, these approaches require a substantial amount of historical data for training and necessitate the collection of extensive data whenever there is a change in the grid. Conversely, the \acrshort{drl} based approach aims to gradually learn how to optimize power flow in electrical networks and dynamically identify the optimal operating point. Approaches like \cite{OPF_RL2}, \cite{opf_drl}, \cite{opf_rl3}, and \cite{opf_ppo} utilize different \acrshort{drl} algorithms to solve the \acrshort{opf}, but none incorporate \acrshort{gnn}, resulting in a loss of generalization capability.

\section{\uppercase{Problem statement}}
\label{sec:problem_statement}
\begin{figure*}[h]
    \centering
    \includegraphics[width=0.8\textwidth]{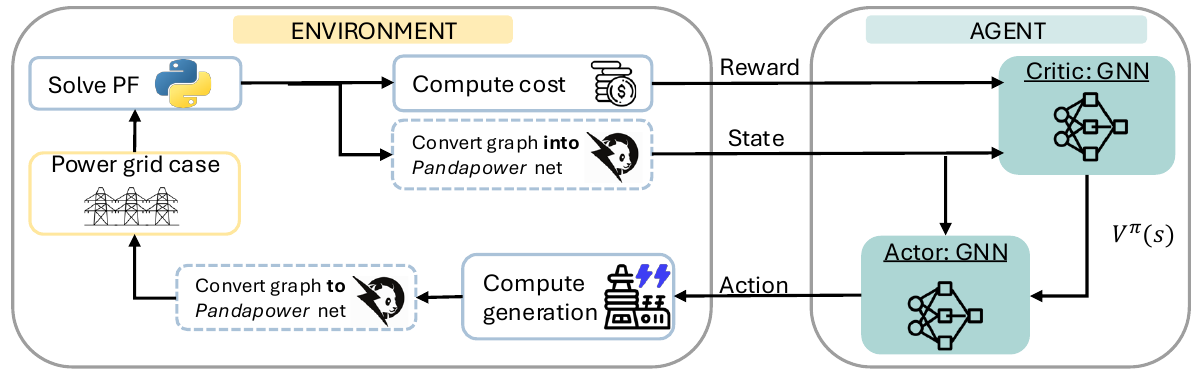}
    \caption{Overview of the PPO-based architecture for power grid optimization. The system consists of an environment and an agent. The environment simulates a power grid case. The agent, implemented using PPO with GNN, consists of an actor-critic structure: the Actor-GNN selects actions, while the Critic-GNN evaluates state values. The agent interacts with the environment by receiving state information, actions (change generation), and rewards based on the computed cost.}
    \label{ppo_diagrama2}
\end{figure*} 
As illustrated in \autoref{ppo_diagrama2}, an agent is trained using \acrshort{drl}. Over multiple timesteps, the agent iteratively modifies the generation values of a power grid, aiming to maximize the reward. This reward reflects the reduction in generation cost compared to the previous timestep. The training process begins with a base case where the agent minimizes the cost while optimizing the search for feasible solutions. Once trained, this agent can be deployed to compute \acrshort{opf} in power grids with altered topologies, such as the loss of an electrical line due to maintenance or the disconnection of a load. The results obtained, in terms of cost, are often better or comparable to those achieved using \acrshort{dcopf}.

\section{\uppercase{background}}
\label{sec:background}
In this section, we provide the necessary background for \acrshort{gnn} (\autoref{subsec:gnn}), the \acrshort{drl} algorithm used (\autoref{subsec:drl}) and we expand the definition of \acrshort{opf}. Commonly, \acrshort{opf} minimizes the generation cost, so the objective is to minimize the cost of power generation while satisfying operating constraints and demands. Some of these constraints are restrictions of both maximum and minimum voltage in the nodes or that the net power in each bus is equal to the power consumed minus generated \cite{pf_hist}. At the same time, the \acrfull{pf} or load flow refers to the generation, load, and transmission network equations. It is a quantitative study to determine the flow of electric power in a network under given load conditions whose objective is to determine the steady-state operating values of an electrical network \cite{pf_hist}.

\subsection{Graph Neural Networks}
\label{subsec:gnn}
\acrlong{gnn} are methods based on deep learning that function within the graph domain. Due to their effectiveness, \acrshort{gnn} has recently emerged as a widely utilized approach for graph analysis \cite{gnn_review}. The concept of \acrshort{gnn} was first introduced by \cite{gnn1}. This architecture can be viewed as a generalization of convolutional neural networks tailored for graph structures, achieved by unfolding a finite number of iterations. We employ \acrfull{mpnn}, as introduced in \cite{gnn_chemistry}, which represents a specific type of \acrshort{gnn} that operates through an iterative message-passing algorithm, facilitating the propagation of information among elements in a graph $G=(N, E)$. Initially, the hidden states of the nodes are set using the graph’s node-level features from the data. Subsequently, the message-passing process unfolds \cite{gnn_chemistry}: Message (\autoref{gnn_1}), Aggregation (\autoref{gnn_1}), and Update (\autoref{gnn_2}). After a defined number of message-passing steps, a readout function $r(\cdot)$ takes the final node states $h_{v}^{K}$ as input to generate the ultimate output of the \acrshort{gnn} model. The readout can predict various outcomes at different levels, depending on the specific problem at hand.

\small
\begin{gather}
    M_{v}^{k} = a(\{m(h_{v}^{k}, h_{k}^{i} )\}_{i \in \beta(v)})\label{gnn_1}\\
    h_{v}^{k+1} = u(h_{v}^{k}, M_{k}^{v} )\label{gnn_2}
\end{gather} 
\normalsize

\subsection{Deep Reinforcement Learning}
\label{subsec:drl}
The objective in \acrfull{rl} is to learn a behavior (policy). In \acrshort{rl}, an agent acquires a behavior through interaction with an environment to achieve a specific goal \cite{ppo_original}. This approach is grounded in the reward assumption: all objectives can be framed as the result of maximizing cumulative rewards. \acrshort{drl} is recognized for its robust ability to tackle complex decision-making challenges, making it suitable for capturing the dynamics involved in the power flow reallocation process \cite{opf_drl}.

Within the \acrshort{drl} algorithms, we use \acrlong{ppo}, formulated in 2017 \cite{ppo_original} and becoming the default reinforcement learning algorithm at \textit{OpenAI} \cite{ppo_original} because of its ease of use and its good performance. As an actor-critic algorithm, the critic evaluates the current policy and the result is used in the policy training. The actor implements the policy and it is trained using Policy Gradient with estimations from the critic \cite{ppo_original}. \acrshort{ppo} strikes a balance between ease of implementation, sample complexity, and ease of tuning, trying to compute an update at each step that minimizes the cost function while ensuring that the deviation from the previous policy is relatively small \cite{ppo_original}. \acrshort{ppo} uses Trust Region and imposes policy ratio to stay within a small interval (policy ratio $r_{t}$ is clipped), $rt$ will only grow to as much as $1 + \varepsilon$ (\autoref{clip1}) \cite{ppo_original}. The total loss function for the \acrshort{ppo} comprises $L^{CLIP}$ (\autoref{clip1}), the mean-square error loss of the value estimator (critic loss), and an additional term that promotes higher entropy (enhancing exploration) (\autoref{clip2}). \acrshort{ppo} employs \acrfull{gae} to compute the advantage ($\hat{A}_{t}$), as shown in \autoref{gae2}. This advantage method is detailed in \cite{gae_paper}.

\small
\begin{gather}
    L_{TOTAL} = L^{CLIP} + L^{VALUE}*k_{1}-L^{ENTROPY}*k2 \label{clip2}\\
    L^{CLIP} (\theta) = \hat{E_{t}} \left[ min (r_{t}(\theta) \hat{A}_{t}, clip(r_{t}(\theta), 1 - \epsilon, 1 + \epsilon) \hat{A}_{t}) \right] \label{clip1}\\
    A^{GAE}_{0} = \delta_{0} + (\lambda \gamma) A^{GAE}_{1} \label{gae2}
\end{gather}
\normalsize

\section{\uppercase{proposed method}}
\label{sec:methodology}

In this section, we outline our approach, which is schematically illustrated in \autoref{ppo_diagrama2}. Both the actor and critic of the \acrshort{drl} agent are represented as \acrshort{gnn}, while the state of the environment corresponds to the resulting graph of the power grid. Within the \acrshort{drl} environment, the agent executes an action at each time step, adjusting the power of the generator. Subsequently, the power grid graph is updated through a \acrlong{pf}.

We treat our power grid as graph-structured data by utilizing information on the power grid topology, where electrical lines serve as edges and buses as nodes, along with the associated loads and generations. For the electrical lines, we define features using resistance $R$ and reactance $X$ ($e_{n,n}^{ACLine} = [ R_{n,m}, X_{n,m}]$). For the buses, we incorporate voltage information, including its magnitude $V$ and phase angle $\theta$, as well as the power exchanged at that bus between the connected loads and generators, represented as $X_{n}^{AC} = [V_{n}, \theta_{n}, P_{n}, Q_{n}]$. 

\begin{figure*}[t]
    \centering
    \includegraphics[width=0.9\textwidth]{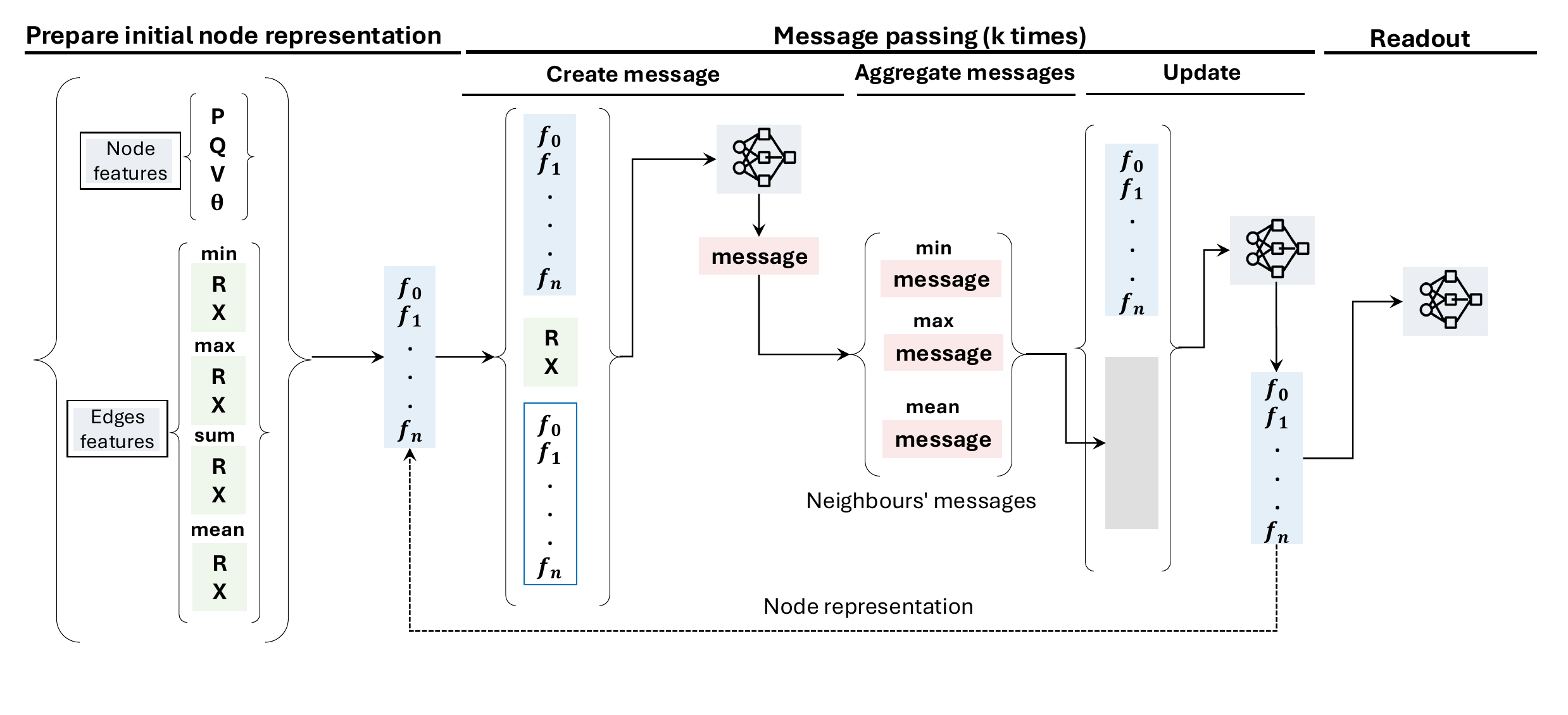}
    \caption{\acrshort{gnn} architecture. Both critic and actor employ the same \acrshort{gnn} architecture, differing only in their readout layers. The process initiates with the preparation of initial node representations, leveraging both node and edge features. Specifically, the GNN's input comprises the electrical parameters of the grid. During the message-passing phase (repeated k times), each node generates messages based on its features, which are subsequently aggregated from its neighbors. These aggregated messages refine the node representations via an update function. Ultimately, in the readout phase, the actor utilizes these refined representations to compute the action, while the critic uses them to estimate the value function.}
    \label{gnn_arq}
\end{figure*}

The overall architecture of the \acrshort{gnn} is illustrated in \autoref{gnn_arq}, which includes the message passing and readout components. At each message-passing step $k$, each node $v$ receives the current hidden states of all nodes in its neighborhood and processes them individually by applying a message function $m(·)$ (NN) along with its own internal state $h_{v}^{k}$ and the features of the connecting edge. These messages are then aggregated through a concatenation of min, max, and mean operations. By combining this message aggregation with the node's hidden state and updating the combination using another NN, new hidden state representations are generated. After a specified number of message passing steps, a readout function $r(·)$ takes the final node states $h_{v}^{K}$ as input to produce the final output of the GNN model.

For the \textbf{actor}, whose output is the \acrshort{rl} policy, the readout consists of a 3-layer MLP \acrshort{nn} where the input comprises each of the \textit{node representations}. We independently pass through this readout the representation of all nodes with a generator, resulting in $N$ output values. Each output value signifies the probability of selecting that generator to enhance its power. This approach to managing the readout ensures that the architecture remains generalizable to any number of generators. These values are utilized to form a probability distribution, from which a value is sampled (representing the ID of the generator whose generation is increased at that time horizon $t$). The \textbf{critic} employs a centralized readout that takes all node hidden states as inputs (by concatenating the sum, minimum, and maximum), producing an output that estimates the value function. Consequently, the input dimension is \textit{3*node representation} with a single output for the entire graph. The critic is also structured as a 3-layer MLP.
 
Regarding the environment with which the agent interacts, at each time instant $t$, the state is defined by the graph updated by the \acrlong{pf}. In each horizon step ($t$), the action performed by the agent involves increasing the generation of one of the generator nodes. The agent will determine which of the available generators will have its generation increased by one portion. For each generator, the power range between its maximum and minimum power is divided into $N$ portions. When a generator reaches its maximum power, generation cannot be increased, resulting in the power grid (and thus the state of the environment) remaining unchanged. If the \acrshort{pf} does not converge, it indicates that with the given demand and generation, meeting the constraints is not feasible; we refer to this situation as an infeasible solution. When initializing an episode (the initial state of the power grid), we aim for the generation to be as low as possible, allowing the agent to raise it until it reaches the optimum. Additionally, we must consider that if the generation is too low in the initial time steps, the solution may become infeasible. Consequently, we decide to set the minimum generation at 20\%. The reward at time $t$ is calculated as the improvement in the solution's cost compared to $t-1$, as shown in \autoref{reward_eq}. The reward is positive during a time step when the agent's action results in a decrease in generation cost. Conversely, if the agent selects a generator that is already at its maximum capacity, leads to an infeasible solution, or increases the cost, the reward will be negative.

\scriptsize
\begin{equation}
    r(t)=\left\lbrace\begin{array}{l} 
    MinMaxScaler(cost) - Last(MinMaxScaler(cost))
    \\ cte_{1} \text{       if selected generator already in } P_{max}
    \\ cte_{2} \text{       if solution no feasible}
    \end{array}\right.
    \label{reward_eq}
\end{equation} 
\normalsize

\section{\uppercase{Performance evaluation}} 
\label{sec:evaluation}
This section outlines the experimentation conducted to validate the proposed approach, the data utilized, and discusses the results obtained. 

\textbf{Overview:} We train the agent using a base case and subsequently evaluate its performance in modified scenarios. On one hand, we adjust the number of loads and their values, as real power grid operations involve continuous changes in loads. On the other hand, we simulate the unavailability of certain electrical lines due to breakdowns or maintenance. This approach demonstrates that the agent, once trained, can generalize to previously unseen cases. We compare the cost differences between our method and the industry standard method, the \acrshort{dcopf}. Our goal is to demonstrate that our method can produce a solution that is equal to or better than the \acrshort{dcopf}, while avoiding its disadvantages.

\subsection{Experimental Setting}

\begin{figure}[htb]
    \centering
    \includegraphics[width=0.49\textwidth]{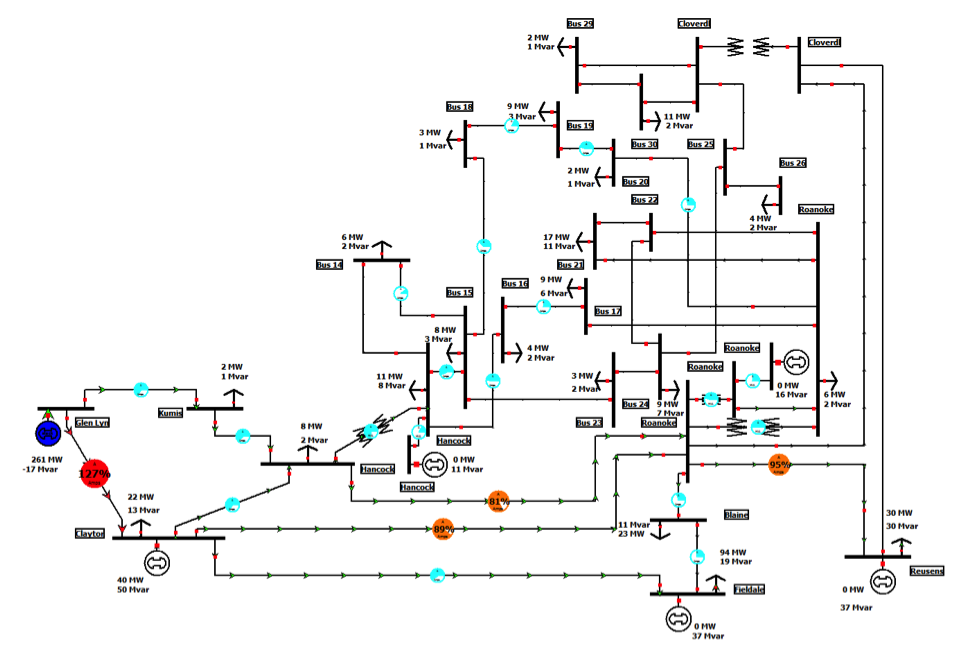}
    \caption{IEEE 30 bus system \cite{pandapower2}.}
    \label{ieee30dataset}
\end{figure}

We train the agent using the IEEE 30 bus system as our case study (\autoref{ieee30dataset}). This system consists of thirty nodes, forty links, five generators, and twenty loads, with all generators modeled as thermal generators. We utilize \textit{Pandapower}, a Python-based, BSD-licensed power system analysis tool \cite{pandapower}. This tool enables us to perform calculations such as \acrshort{opf}  using the IPOPT optimizer and \acrshort{pf} analysis, which we employ to evaluate our costs and update our environment. Additionally, this library allows us to verify the physical feasibility of our solutions by ensuring they comply with \acrshort{pf} constraints.

The objective of the training is to optimize the parameters so that the actor becomes a good estimator of the optimal global policy and the critic learns to approximate the state value function of any global state. Many hyperparameters can be modified, and they are divided into different groups. Grid search has been performed on many of them, and the final selected values of the most important ones are shown below. 
\begin{itemize}
    \item Related to learning loop: Minibatch (25), epochs (3) and optimizer (ADAM), with its parameters like learning rate $lr$ (0.003).
    \item Related to the power grid: Generator portions (50).
    \item Related to \acrshort{rl}: Episodes (500), horizon size $T$ (125), reward $cte_{1}$ (-1) and $cte_{2}$ (-2).
    \item Actor and Critic \acrshort{gnn}: Message iterations $k$ (4), node representation size (16). The NN to create the messages is a 2-layer MLP and the updated one is 3-layer MLP.
\end{itemize}
\acrshort{ppo} is an online algorithm that, similar to other reinforcement learning algorithms, learns from experience. The training pipeline is organized as follows:
\begin{itemize}
    \item An episode of length $T$ is generated by following the current policy. While at the same time the critic’s value function $V$ evaluates each visited global state; this defines a trajectory $\left\lbrace s_{t}, a_{t}, r_{t}, p_{t}, V_{t}, s_{t+1}\right\rbrace_{t=0}^{T-1}$.
    \item This trajectory is used to update the model parameters –through several epochs of minibatch Stochastic Gradient Descent– by maximizing the global \acrshort{ppo} objective.
\end{itemize}
The same process of generating episodes and updating the model is repeated for a fixed number of iterations to guarantee convergence. \textit{MinMaxScaler} has been used for data preprocessing for node features, edge features and generation output. More implementation details can be found in the public repository \footnote{\url{https://github.com/anlopez94/opf_gnn_ppo}}. 

\subsection{Experimental Results}

Once the training has been done and the best combination of hyperparameters, network design and reward modelling has been chosen, the best checkpoint of the model is selected to compute \acrshort{opf} in different networks. To validate our solution, we use the deviation of the cost concerning the minimum cost, the one obtained with the \acrshort{acopf} (\%DRL+OPF perf. in \autoref{test_load1} - \autoref{test_edges}). We compare it with the cost deviation obtained with \acrshort{dcopf} (\%DCOPF perf in \autoref{test_load1} - \autoref{test_edges}). We compute the ratio between these two deviations. We calculate the improvement ratio by dividing the first value by the second, which reflects the enhancement over the \acrshort{dcopf}.

Once the model is trained, only the actor part is used in the evaluation. During $T$ steps of an episode, the actions sampled from the probability distribution obtained from the actor for each state of the network are executed. Finally, the mean cost of the best ten evaluations is measured, as well as the convergence of the problem. We evaluate $100$ times for each test case. In \autoref{test_load1} - \autoref{test_edges}, it is highlighted between the deviation in \% of our solution concerning the \acrshort{opf}'s one and the deviation obtained by the \acrshort{dcopf}. We also assess the convergence and physical feasibility of our solution, finding that it was feasible in the majority of cases.

First, all network loads are varied by multiplying their value by a random number between a value less than 1 and a value higher than 1 (\autoref{test_load1}). Each row in the table is a test in which the name specifies the upper and lower percentages by which loads have been varied. In all tests, performance with our method is better with ratios of up to 1.30.

\begin{table}[htbp]
  \centering
  \caption{Results on case IEEE 30 varying loads from base case.}
  \begin{adjustbox}{width=0.45\textwidth}
    \begin{tabular}{cccc}
    \hline
    {} &  \% DRL+OPF perf. &  \%DCOPF perf. &  ratio \\
    \hline
    load\_inf0.1\_sup0.1 & \textbf{0,75} & 0,77 & 1,02  \\
    load\_inf0.2\_sup0.1 & \textbf{0,59} & 0,68 & 1,16 \\
    load\_inf0.3\_sup0.1 & \textbf{0,53} & 0,73 & 1,38 \\
    load\_inf0.4\_sup0.1 & \textbf{0,61} & 0,67 & 1,10 \\
    \hline
    \end{tabular}
 \end{adjustbox}
  
  \label{test_load1}
\end{table}

After varying the load value, we experiment with removing $n$ loads from the grid. We randomly choose several loads, remove them from the network and evaluate the model (\autoref{test_loads2}). Each row in the table is a test in which the name specifies the number of loads that have been removed. Our cost deviation is lower or similar than the \acrshort{dcopf} even by eliminating almost 50\% of the loads. Finally, in \autoref{test_edges} we show the results of creating networks from the original one by removing one or more electrical lines (edges). Each row in the table is a test in which the name specifies the number of power lines removed.

\begin{table}[htbp]
  \centering
  \caption{Results on case IEEE 30 removing loads from base case.}
  \begin{adjustbox}{width=0.45\textwidth}
    \begin{tabular}{cccc}
    \hline
    {} &  \% DRL+OPF perf. &  \%DCOPF perf. &  ratio \\
    \hline
    load\_1 & \textbf{0,67} & 0,72 & 1,07 \\
    load\_2 & \textbf{0,71} & 0,71 & 1,00 \\
    load\_3 & \textbf{0,67} & 0,67 & 1,00 \\
    load\_4 & 1,06 & \textbf{0,68} & 0,65 \\
    load\_5 & \textbf{0,61} & 0,64 & 1,05 \\
    load\_8 & 1,10 & \textbf{0,63} & 0,52 \\
    \hline
    \end{tabular}
 \end{adjustbox}
  
  \label{test_loads2}
\end{table}

\begin{table}[htbp]
  \centering
    \caption{Results on case IEEE 30 removing edges from base case.}
    \begin{adjustbox}{width=0.45\textwidth}
    \begin{tabular}{cccc}
    \hline
    {} &  \% DRL+OPF perf. &  \%DCOPF perf. &  ratio \\
    \hline
   edge\_1 & \textbf{0,73} & 0,77 & 1,05 \\
   edge\_2 & \textbf{0,41} & 1,16 & 2,83 \\
   edge\_3 & 0,62 & \textbf{0,61} & 0,99 \\
   edge\_4 & \textbf{0,65} & 0,88 & 1,36 \\
   edge\_5 & \textbf{0,90} & 0,96 & 1,07 \\
   edge\_8 & \textbf{0,59} & 0,89 & 1,51 \\
    \hline
    \end{tabular}
 \end{adjustbox}
  \label{test_edges}
\end{table}

In experiments removing electrical lines (\autoref{test_edges}) as more power lines are removed (more than 8), sometimes, the agent does not find a good feasible solution (no convergence). When we experimented with changing the load values in the second test (\autoref{test_load1}), we observed that increasing the loads by more than 10\% caused the tests to fail to converge. With the other changes in topology, 100\% of tests converged, so we can conclude that our model is capable of generalizing to unseen topologies (based on the trained one).

\section{\uppercase{discussion}}
\label{sec:discussion}

We have successfully designed a solution to address the \acrshort{opf}, capable of generalization, utilizing \acrshort{drl} and \acrshort{gnn}. The network topology has been modified, and we have demonstrated that the agent can identify a strong solution (with performance closely aligned to the current industry standard \acrshort{dcopf}), ensuring that this solution is both feasible and compliant with the constraints. Thanks to the design of \acrshort{gnn}, it can be trained on various cases and subsequently applied to different scenarios. In this paper, we validate that this architecture effectively tackles \acrshort{opf}, showcasing the generalization capability of our solution by considering modifications to the network scenario encountered during training (including different loads and a reduction in the number of edges). By integrating these two technologies for the first time, we conclude that their combination is feasible, leveraging the advantages of both. Our findings indicate that the proposed architecture represents a promising initial step toward solving the \acrshort{opf}. Future work could explore the incorporation of additional features in the node representation, such as the maximum and minimum allowable voltage and model other types of electrical generation.
\section*{Acknowledgments}
This research is supported by the Industrial Doctorate Plan of the Department of Research and Universities of the Generalitat de Catalunya, under Grant AGAUR 2023 DI060.
\bibliographystyle{apalike}
\bibliography{bibliography}

\begin{thebibliography}{}

\bibitem[Cao et~al., 2021]{opf_rl3}
Cao, D., Hu, W., Xu, X., Wu, Q., Huang, Q., Chen, Z., and Blaabjerg, F. (2021).
\newblock Deep reinforcement learning based approach for optimal power flow of distribution networks embedded with renewable energy and storage devices.
\newblock {\em Journal of Modern Power Systems and Clean Energy}, 9(5):1101--1110.

\bibitem[Diehl, 2019]{opf_gnn_cutre}
Diehl, F. (2019).
\newblock Warm-starting ac optimal power flow with graph neural networks.
\newblock In {\em 33rd Conference on Neural Information Processing Systems (NeurIPS 2019)}, pages 1--6.

\bibitem[Donnot et~al., 2017]{opf_ml2}
Donnot, B., Guyon, I.~M., Schoenauer, M., Marot, A., and Panciatici, P. (2017).
\newblock Fast power system security analysis with guided dropout.
\newblock {\em ArXiv}, abs/1801.09870.

\bibitem[Donon et~al., 2020]{balthazar2}
Donon, B., Clément, R., Donnot, B., Marot, A., Guyon, I., and Schoenauer, M. (2020).
\newblock Neural networks for power flow: Graph neural solver.
\newblock In {\em Electric Power Systems Research}, volume 189, page 106547.

\bibitem[Donon et~al., 2019]{balthazar1}
Donon, B., Donnot, B., Guyon, I., and Marot, A. (2019).
\newblock {Graph Neural Solver for Power Systems}.
\newblock In {\em {IJCNN 2019 - International Joint Conference on Neural Networks}}, Budapest, Hungary.

\bibitem[Fraunhofer, 2022]{pandapower2}
Fraunhofer, IEE, U. o.~K. (2022).
\newblock {Pandapower} documentation.

\bibitem[Gilmer et~al., 2017]{gnn_chemistry}
Gilmer, J., Schoenholz, S.~S., Riley, P.~F., Vinyals, O., and Dahl, G.~E. (2017).
\newblock Neural message passing for quantum chemistry.
\newblock In {\em International conference on machine learning}, pages 1263--1272. PMLR.

\bibitem[Li et~al., 2021]{opf_drl}
Li, J., Zhang, R., Wang, H., Liu, Z., Lai, H., and Zhang, Y. (2021).
\newblock Deep reinforcement learning for optimal power flow with renewables using graph information.
\newblock {\em ArXiv}, abs/2112.11461.

\bibitem[Liao et~al., 2022]{review_gnn_power}
Liao, W., Bak-Jensen, B., Pillai, J.~R., Wang, Y., and Wang, Y. (2022).
\newblock A review of graph neural networks and their applications in power systems.
\newblock {\em Journal of Modern Power Systems and Clean Energy}, 10(2):345--360.

\bibitem[Mary et~al., 2012]{pf_hist}
Mary, A., Cain, B., and O’Neill, R. (2012).
\newblock History of optimal power flow and formulations.
\newblock {\em Fed. Energy Regul. Comm.}, 1:1--36.

\bibitem[Munikoti et~al., 2024]{drl_gnn_rev}
Munikoti, S., Agarwal, D., Das, L., Halappanavar, M., and Natarajan, B. (2024).
\newblock Challenges and opportunities in deep reinforcement learning with graph neural networks: A comprehensive review of algorithms and applications.
\newblock {\em IEEE Transactions on Neural Networks and Learning Systems}, 35(11):15051--15071.

\bibitem[Owerko et~al., 2020]{rivero1}
Owerko, D., Gama, F., and Ribeiro, A. (2020).
\newblock Optimal power flow using graph neural networks.
\newblock In {\em ICASSP 2020 - 2020 IEEE International Conference on Acoustics, Speech and Signal Processing (ICASSP)}, pages 5930--5934.

\bibitem[Pan et~al., 2022]{deep_opf}
Pan, X., Chen, M., Zhao, T., and Low, S.~H. (2022).
\newblock Deepopf: A feasibility-optimized deep neural network approach for ac optimal power flow problems.
\newblock In {\em IEEE Systems Journal}, pages 1--11.

\bibitem[Scarselli et~al., 2009]{gnn1}
Scarselli, F., Gori, M., Tsoi, A.~C., Hagenbuchner, M., and Monfardini, G. (2009).
\newblock The graph neural network model.
\newblock {\em IEEE Transactions on Neural Networks}, 20:61--80.

\bibitem[Schulman et~al., 2015]{gae_paper}
Schulman, J., Moritz, P., Levine, S., Jordan, M., and Abbeel, P. (2015).
\newblock High-dimensional continuous control using generalized advantage estimation.
\newblock {\em ArXiv}, abs/1506.02438.

\bibitem[Schulman et~al., 2017]{ppo_original}
Schulman, J., Wolski, F., Dhariwal, P., Radford, A., and Klimov, O. (2017).
\newblock Proximal policy optimization algorithms.
\newblock {\em ArXiv}, abs/1707.06347.

\bibitem[Thurner et~al., 2018]{pandapower}
Thurner, L., Scheidler, A., Schäfer, F., Menke, J.-H., Dollichon, J., Meier, F., Meinecke, S., and Braun, M. (2018).
\newblock Pandapower—an open-source python tool for convenient modeling, analysis, and optimization of electric power systems.
\newblock {\em IEEE Transactions on Power Systems}, 33(6):6510--6521.

\bibitem[Wood et~al., 2013]{elec_book}
Wood, A.~J., Wollenberg, B.~F., and Sheble, G.~B. (2013).
\newblock {\em Power Generation, Operation, and Control}.
\newblock John Wiley \& Sons, Hoboken, NJ, USA, 3rd edition.

\bibitem[Zhen et~al., 2022]{OPF_RL2}
Zhen, H., Zhai, H., Ma, W., Zhao, L., Weng, Y., Xu, Y., Shi, J., and He, X. (2022).
\newblock Design and tests of reinforcement-learning-based optimal power flow solution generator.
\newblock {\em Energy Reports}, 8:43--50.
\newblock 2021 The 8th International Conference on Power and Energy Systems Engineering.

\bibitem[Zhou et~al., 2020a]{gnn_review}
Zhou, J., Cui, G., Hu, S., Zhang, Z., Yang, C., Liu, Z., Wang, L., Li, C., and Sun, M. (2020a).
\newblock Graph neural networks: A review of methods and applications.
\newblock {\em AI Open}, 1:57--81.

\bibitem[Zhou et~al., 2020b]{opf_ppo}
Zhou, Y., Zhang, B., Xu, C., Lan, T., Diao, R., Shi, D., Wang, Z., and Lee, W.-J. (2020b).
\newblock A data-driven method for fast ac optimal power flow solutions via deep reinforcement learning.
\newblock {\em Journal of Modern Power Systems and Clean Energy}, 8(6):1128--1139.

\end{thebibliography}

\end{document}